\documentclass{article}
\usepackage{amsmath}
\usepackage{graphicx}
\usepackage{url}
\usepackage{authblk}
\usepackage{relsize}
\usepackage{xcolor}

\let\citeleft=(
\let\citeright=)

\usepackage[normalem]{ulem}

\newcommand{\revised}[1]{%
\ifx\highlightrevisions\undefined{#1}%
\else\textcolor{red}{#1}%
\fi}

\newcommand{\reviewercommentnum}[1]{%
\ifx\showreviewercommentnum\undefined%
\else{[\bf{#1}] }%
\fi}

\newcommand{\erased}[1]{%
\ifx\showerased\undefined%
\else{\sout{#1}}%
\fi}

\title{Estimating Absolute-Phase Maps Using ESPIRiT and Virtual Conjugate Coils}
\author{Martin Uecker$^1$ and Michael Lustig$^2$}

\affil{
$^1$Institute for Diagnostic and Interventional Radiology, \\
    University Medical Center G\"ottingen, G\"ottingen, Germany. \\
$^2$German Centre for Cardiovascular Research (DZHK) \\
$^3$Department of Electrical Engineering and Computer Sciences, \\
University of California, Berkeley, California.}

\begin{document}

\maketitle

\vfill

\noindent
{\em Running head:} Estimating Absolute-Phase Maps Using ESPIRiT

\vspace{0.5cm}

\noindent
{\em Address correspondence to:} \\
    Martin Uecker \\
    University Medical Center G\"ottingen \\
    Institute for Diagnostic and Interventional Radiology \\
    Robert-Koch-Str. 40\\
    37075 G\"ottingen, Germany \\
    martin.uecker@med.uni-goettingen.de

\vspace{0.5cm}
\noindent
{Supported by: NIH R01EB019241, NIH R01EB009690, NIH P41RR09784, Sloan Research Fellowship, Okawa Research grant, and GE Healthcare.}

\vspace{0.3cm}
\noindent
Approximate word count: {197 (Abstract) 2780 (body)

\vspace{0.3cm}
\noindent
Submitted to {\it Magnetic Resonance in Medicine} as a Note.

\vspace{0.3cm}
\noindent
Part of this work has been presented at the ISMRM Annual Conference 2014.

\newpage

\section*{Abstract}
{\bf Purpose:} To develop an ESPIRiT-based method to estimate coil
sensitivities with image phase as a building block
for efficient and robust image reconstruction with phase constraints.
{\bf Theory and Methods:}
ESPIRiT is a new framework for calibration of the coil sensitivities 
and reconstruction in parallel Magnetic Resonance Imaging (MRI).
Applying ESPIRiT to a combined set of physical and virtual conjugate 
coils (VCC-ESPIRiT) implicitly exploits
conjugate symmetry in k-space similar to VCC-GRAPPA. Based on this
method, a new post-processing step is proposed for the explicit computation 
of  coil sensitivities that include the absolute phase of the image.
The accuracy of the  computed maps is directly
validated using a test based on projection onto fully sampled coil images
and also indirectly in phase-constrained parallel-imaging 
reconstructions.
{\bf Results:}
The proposed method can estimate accurate sensitivities which
include low-resolution image phase. In case of high-frequency phase 
variations VCC-ESPIRiT yields an additional set of maps that
indicates the existence of a high-frequency phase component. Taking
this additional set of maps into account can improve the robustness
of phase-constrained parallel imaging.
{\bf Conclusion:}
The extended VCC-ESPIRiT is a
useful tool for phase-constrained imaging.

\vspace{0.5in}
\setlength{\parindent}{0in}
{\bf Key words: parallel imaging, partial Fourier, ESPIRiT, virtual coil}
\newpage

\section{Introduction}

The primary quantity measured in MRI, the spin density, can be described by
a real and positive function. In principle, this prior knowledge can
reduce the amount of k-space data necessary to reconstruct an image to one
half~\cite{Margosian1986}. In practice, various phase effects from B1, flow, 
off-resonance, and others cause phase variations in the image. 
By using a low-resolution phase map, homodyne reconstruction~\cite{HOMODYNE} 
or SENSE-based parallel imaging with a phase constraint can sometimes
be applied~\cite{PruessmannSENSE,Willig2005}. Especially
in the later case, this causes reconstruction errors if the data is
affected by phase variations that are not described
by the low-resolution phase map.
Recently, ESPIRiT has
been described - a new method to obtain highly accurate estimations of the
coil sensitivities from a fully-sampled calibration region in the k-space 
center~\cite{ESPIRiT}.
The estimates of the maps are defined up to multiplication with an unknown
complex-valued function, {\it i.e.} only relative coil sensitivities are obtained.
Here, we demonstrate an extension to ESPIRiT that can be used
to estimate coil sensitivities with phase that make the reconstructed image
real-valued with high accuracy. This is accomplished by applying ESPIRiT to
virtual conjugate coils (VCC-ESPIRiT), which have been introduced previously to improve
GRAPPA reconstruction~\cite{VCGRAPPA,Blaimer2015}. 
Instead of doing a reconstruction using virtual conjugate
coils which implicitly exploits conjugate symmetry, the new post-processing
technique estimates sensitivities with image phase. The reconstruction
can then use an explicit real-value constraint as in 
phase-constrained SENSE which is more flexible and
computationally more efficient.

In cases where the image phase is not smooth,
phase-constrained SENSE with low-resolution phase maps
suffers from reconstruction artifacts. 
While phase-constrained SENSE has been shown to be equivalent 
to SENSE with virtual conjugate coils (VCC-SENSE) which suffers 
from the same problem, methods using calibration in k-space such 
as ESPIRiT or GRAPPA applied to virtual conjugate coils are more
robust~\cite{Blaimer2015}.
Here, we explain the improved robustness with the 
appearance of a second eigenvalue in VCC-ESPIRiT, and
relate it to the idea of explicitly replacing the hard 
phase constraint with regularization of the
imaginary component~\cite{Hoge07}.

Part of this work has been presented at the
22nd ISMRM Annual Conference~\cite{UeckerMilano14}.

\section{Theory}

\begin{figure}[h]
\includegraphics[width=\textwidth]{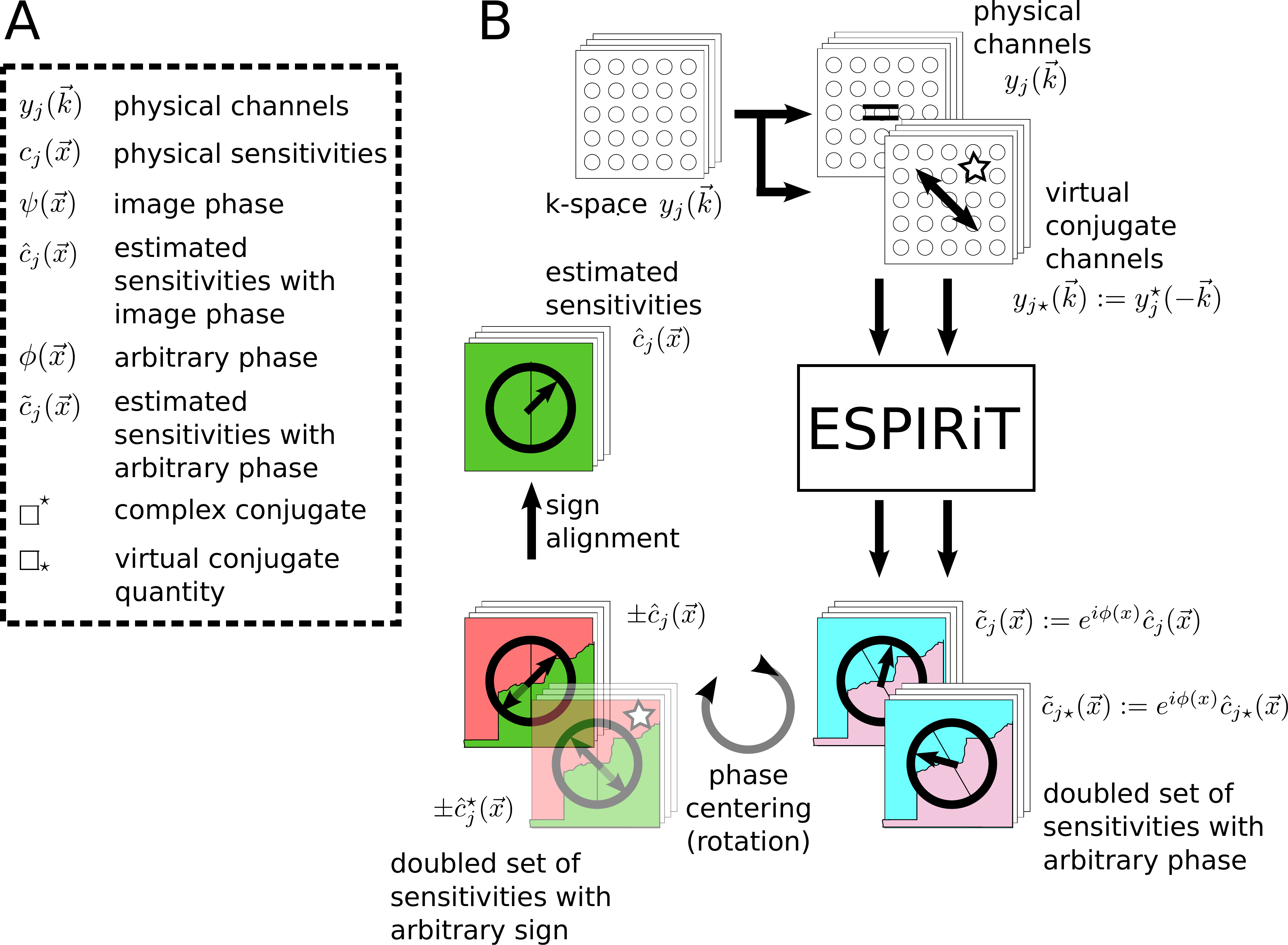}
\caption{\label{FIG1}
{\bf A:} Important symbols used in the text or in the figure.
{\bf B:} Processing steps in the proposed methods (clock-wise): Extension 
of k-space with additional virtual conjugate coils, ESPIRiT calibration,
phase centering, sign unwrapping. After phase centering the 
virtual channels are redundant and can be discarded.}
\end{figure}

Although the spin density is a positive and real quantity the
magnetization image $\rho(\vec x)$ measured in MRI usually has phase.
Formally, the image $\rho(\vec x)$ can be assumed to be real-valued
in the signal equations for parallel imaging  if the image phase 
$\psi(\vec x)$ is fully absorbed into the coil sensitivities:
\begin{align*}
	y_j(\vec k) & = \int \textrm{d}\vec x \, \underbrace{|\rho(\vec x)| e^{i\psi(\vec x)}}_{\rho(\vec x)} c_j(\vec x) e^{-i2\pi \vec k \cdot \vec x}  \\
		& = \int \textrm{d}\vec x \, |\rho(\vec x)| \underbrace{e^{i\psi(\vec x)} c_j(\vec x)}_{\hat c_j(\vec x)} e^{-i2\pi \vec k \cdot \vec x} 
\end{align*}
Here, $y_j(\vec k)$ represents the k-space sample for the $j$-th
channel at k-space position $\vec k$.  $c_j(\vec x)$ is the physical
coil sensitivity map for channel $j$ at image-domain position $\vec x$
and $\hat c_j$ is the corresponding map with image phase.
For example, such maps with image phase
can be estimated directly from the k-space center \cite{McKenzie2002},
using a recently proposed extension of Walsh's method \cite{Walsh2000,Inati2013},
or by non-linear inversion with real-value constraint~\cite{UeckerSeg}.
In this work, we propose an extension to ESPIRiT
for the estimation of highly accurate maps with phase.

ESPIRiT is a new method for auto-calibrating parallel imaging. It
determines the signal space spanned by local patches in k-space 
using singular value decomposition of a calibration matrix constructed from
auto-calibration data. Because there are local correlations
in k-space due to field-of-view (FOV) limitations and correlations 
induced by the receive coils, this space is a small subspace of 
the space of all possible patches. ESPIRiT recovers the sensitivity maps 
of the receive coils as eigenvectors to the eigenvalue one of
a single reconstruction operator, which itself is derived from the
requirement that all k-space patches lie in the signal subspace. 
In general, subspace-based methods are highly robust to many
types of errors, because the estimated subspace automatically 
adapts to correlations in the data.
In ESPIRiT, multiple sets of sensitivity maps may appear as 
eigenvectors to the eigenvalue one in case the
data does not fit the classical SENSE model.
These additional maps can be taken into account in an extended SENSE-like 
reconstruction which is then as robust as traditional k-space methods.
This has been demonstrated for aliasing in the case
of a small FOV, motion corruption, and chemical shift~\cite{ESPIRiT}.

Because the sensitivities are computed as the point-wise eigenvectors of
a reconstruction operator, the point-wise joint phase is 
undefined. Usually, one channel is selected as a reference and
for each pixel the phase of all channels is rotated so that the reference 
has zero phase.
With an extension to ESPIRiT, sensitivities with phase $\hat c_j$ can be estimated,
{\it i.e.} sensitivities that ideally would result in a real image, as long as
noise and other errors are neglected.
This is achieved by exploiting
the property of ESPIRiT that relative phase between different channels is preserved.
The proposed procedure is described in the following and illustrated in Figure~\ref{FIG1}.
By flipping and conjugating k-space virtual conjugate channels
\begin{align*}
	y_{j\star}(\vec k) & := y_j^{\star}(-\vec k)  = \int \textrm{d}\vec x \, |\rho(\vec x)| \hat c_j^{\star}(\vec x) e^{-i2\pi \vec k \cdot \vec x} 
\end{align*}
are constructed which are then included as additional virtual channels -- doubling the
total number of channels.
ESPIRiT calibration is then applied to the extended data set
\begin{align*}
       [ y_1(\vec k), \cdots, y_N(\vec k), y_{1\star}(\vec k), \cdots y_{N\star}(\vec k) ]
\end{align*}
to estimate the coil sensitivities for all channels.
This yields a vector of sensitivity maps
for all physical and virtual sensitivities up to an unknown pixel-wise phase $\phi(\vec x)$:
\begin{align*}
	[ \tilde c_1(\vec x), \cdots, \tilde c_N(\vec x), \tilde c_{1\star}(\vec x), \cdots \tilde c_{N\star}(\vec x) ]
	= e^{i\phi(\vec x)} [ \hat c_1(\vec x), \cdots, \hat c_N(\vec x), \hat c_{1\star}(\vec x), \cdots \hat c_{N\star}(\vec x) ]
\end{align*}
Note that $\phi(\vec x)$ is an arbitrary phase difference between the estimated
sensitivities~$\tilde c_j$ and the sensitivities with image phase $\hat c_j(\vec x) = e^{i\psi(\vec x)} c_j(\vec x)$, {\it i.e.}
$\tilde c_j(\vec x) = e^{i (\phi(\vec x) + \psi(\vec x))} c_j\vec (x)$.
Because conjugate channels $y_{j\star}$ have conjugate sensitivities,
{\it i.e.} $\hat c_{j\star}(\vec x) = \hat c_j^{\star}(\vec x) $,
the unknown phase $\phi$ can be determined up to $\pi$:
\begin{align*}	
	 \frac{1}{2} \operatorname{Im} \log \sum_j \tilde c_j(\vec x) \tilde c_{j\star}(\vec x) 
			= \frac{1}{2} \operatorname{Im} \log e^{i2\phi(\vec x)} \sum_j |c_j(\vec x)|^2  = \phi(\vec x) + l \pi\qquad l \in \mathbf{Z}
\end{align*}
The remaining ambiguity in phase corresponds to unknown sign in the estimated image.
When enforcing a real-value constraint in SENSE or in similar reconstructions,
the sign ambiguity of the estimated sensitivities can be ignored.
In this work, the sign is simply aligned to a low-resolution estimate of the sensitivities.
After the estimation of the phase, the virtual conjugate coils are not needed anymore
and can be discarded.

A more complicated situation arises if the phase of the image is not
smooth, {\it i.e.} it cannot be represented by a single smooth sensitivity map. 
In case of such inconsistencies VCC-ESPIRiT automatically produces a 
second set of sensitivity maps~\cite{UeckerMilano14,Blaimer2015}. 
As shown later,
taking this second map into account will essentially relax the phase constraint
in affected areas of the image, which can prevent artifacts in phase-constrained
reconstructions.

\section{Methods}

Fully-sampled data from a human brain was acquired with 3D FLASH at 3T (TR/TE = 11/4.9 ms)
using a 32-channel head coil and was retrospectively under-sampled with an
acceleration factor of R = 3 in in the first phase-encoding
dimension. For some experiments, an additional 
partial Fourier factor of PF = 5/8 was applied
in the second phase-encoding direction. Except where stated 
differently, experiments used a
calibration region consisting of 24x24 
auto-calibration signal (ACS) lines.

The Berkeley Advanced Reconstruction Toolbox (BART)
was used for calibration and image 
reconstruction~\cite{UeckerToronto15}.\footnote{\url{https://mrirecon.github.io/bart/}}
In the interest of reproducible research, code and data to
reproduce the experiments are made 
available on Github.\footnote{\url{https://github.com/uecker/vcc-espirit/}}

Sensitivity maps were computed using direct estimation
from the k-space center\cite{McKenzie2002}, ESPIRiT, and the
proposed VCC-ESPIRiT method. For both ESPIRiT-based
methods the following
default parameters were used: The kernel size was six and 
the null-space threshold in the first step of the ESPIRiT calibration 
was 0.001 of the maximum singular-value~\cite{ESPIRiT}. For
image reconstruction, maps were weighted with a smooth S-curve transition
between one and 0.85 of the local eigenvalue~\cite{UeckerSaltLake13}.
In this work, the unknown sign is irrelevant
and has simply been aligned to a low-resolution reference 
to avoid visually distracting sign jumps.

The quality of the estimated maps was directly
evaluated in dependence of the kernel size and
the size of the calibration region. For the evaluation, extended versions
of a previously described projection test can be used~\cite{ESPIRiT}:
Coil images~$m_j$ obtained with discrete Fourier transform from 
fully sampled data were projected onto 
the span of normalized sensitivity maps by summing the images after
multiplication with conjugate sensitivities and then multiplying
with the sensitivities again.
\begin{align*}
	(P m)_j(\vec x) = \frac{\hat c_j(\vec x) \sum_{t=1}^N \hat c^{\star}_t(\vec x) m_t(\vec x)}{\sum_{s=1}^N |\hat c_s(\vec x)|^2}
\end{align*}
This operation is also one of the projections
repeatedly applied in POCSENSE~\cite{POCSENSE}. The result of
the projection $(P m)_j$ is then
subtracted from the original coil images to obtain 
error maps
$E_j(\vec x) = m_j(\vec x) - (P m)_j(\vec x)$ where any  
remaining signal different
from noise indicates that the 
maps do not span the data space correctly.
The error maps for all channels can be combined
into a single error map according to the root-sum-of-squares formula $\sqrt{ \sum_j |E_j|^2 }$.
This procedure can be extended to test how
well the maps describe the image phase by including
a projection onto the real part of the image:
\begin{align*}
	(P_R m)_j(\vec x) = \frac{\hat c_j(\vec x) \operatorname{Re}\sum_{t=1}^N \hat c^{\star}_t(\vec x) m_t(\vec x)}{\sum_{s=1}^N |\hat c_s(\vec x)|^2}
\end{align*}
As another extension,
projecting onto the span of multiple sets can simply be achieved
by summing all individual projections.

Iterative reconstruction with a phase constraint was performed
for different sampling schemes and slices.
Reconstructions using one set of VCC-ESPIRiT maps or two sets of 
VCC-ESPIRiT maps were compared with directly estimated
maps~\cite{McKenzie2002}. In addition, regularization of 
the imaginary component instead of a hard phase constraint 
was used~\cite{Hoge07}.
To evaluate the reconstruction error including phase
errors difference images were computed in the following
way: Invidual coil images were computed by multiplying 
the reconstructed images with the sensitivities. 
Then, for each channel the difference to the corresponding
coil image from fully-sampled k-space data was computed.
Finally, the difference images for all channels were 
combined using root-sum-of-squares formula.

\section{Results}

\begin{figure}[h]
\includegraphics[width=\textwidth]{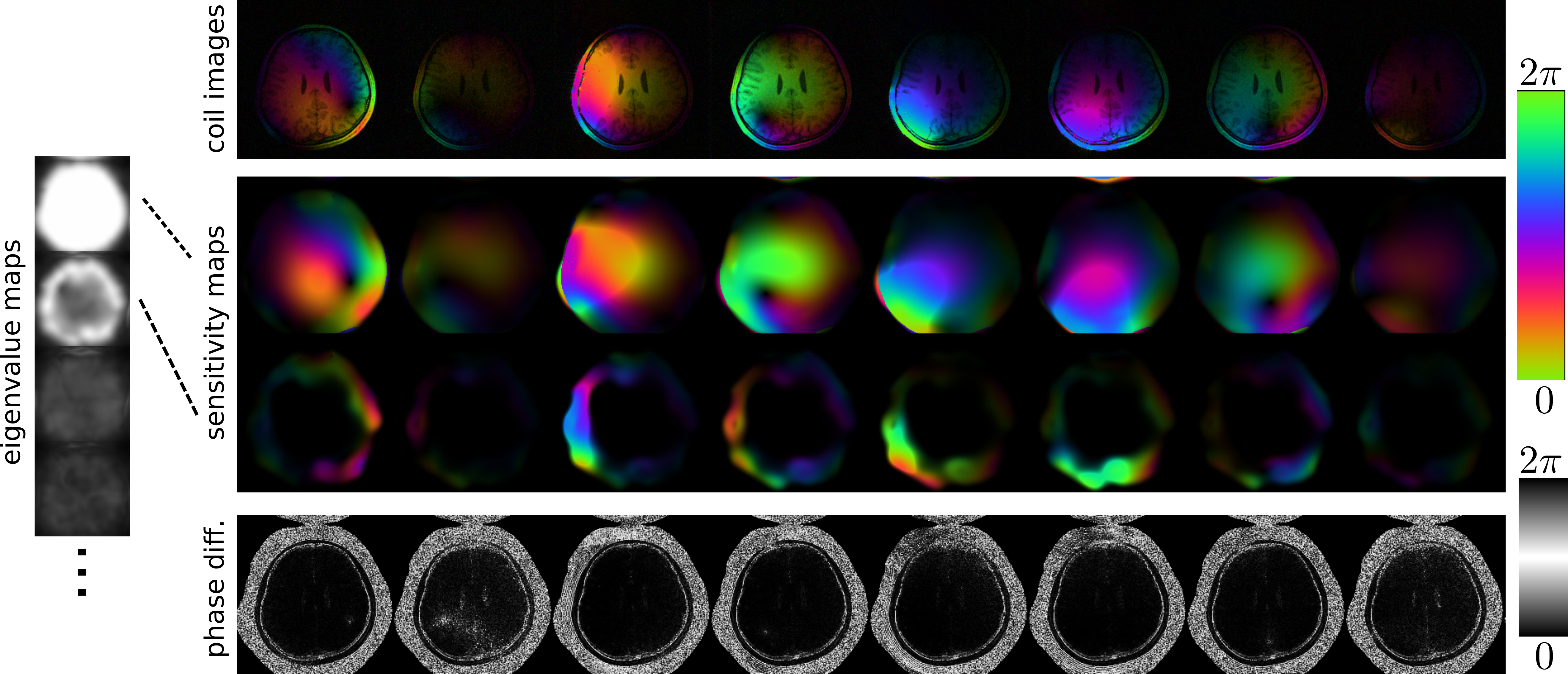}
\caption{\label{FIG2}
VCC-ESPIRiT was applied to a calibration region of 
size 24x24 of a human brain data set.
{\bf Left:}
The first four from 64 eigenvalue maps from VCC-ESPIRiT
calibration. In some regions a second eigenvalue close to 
one indicates the existence of a second set of maps.
{\bf Right:} 
Coil images, the first and second
set of sensitivity (eigenvector) maps, 
and a phase-difference map between the coil images
and the first set of maps are shown (only the first
eight out of 32 physical channels are shown).
Except for the phase-difference map the phase is encoded using color.}
\end{figure}

Figure~\ref{FIG2} shows individual coil images and the first two
sets of maps computed with VCC-ESPIRiT for the first eight channels
of the brain data set. 
The primary set of maps represents the coil sensitivities
with image phase which matches the phase of the 
coil images except for high-frequency phase components. 
A second set of maps appears in image regions affected by this 
high-frequency phase, which - in this example - is caused by 
off-resonance from fat.

\begin{figure}[h]
\includegraphics[width=\textwidth]{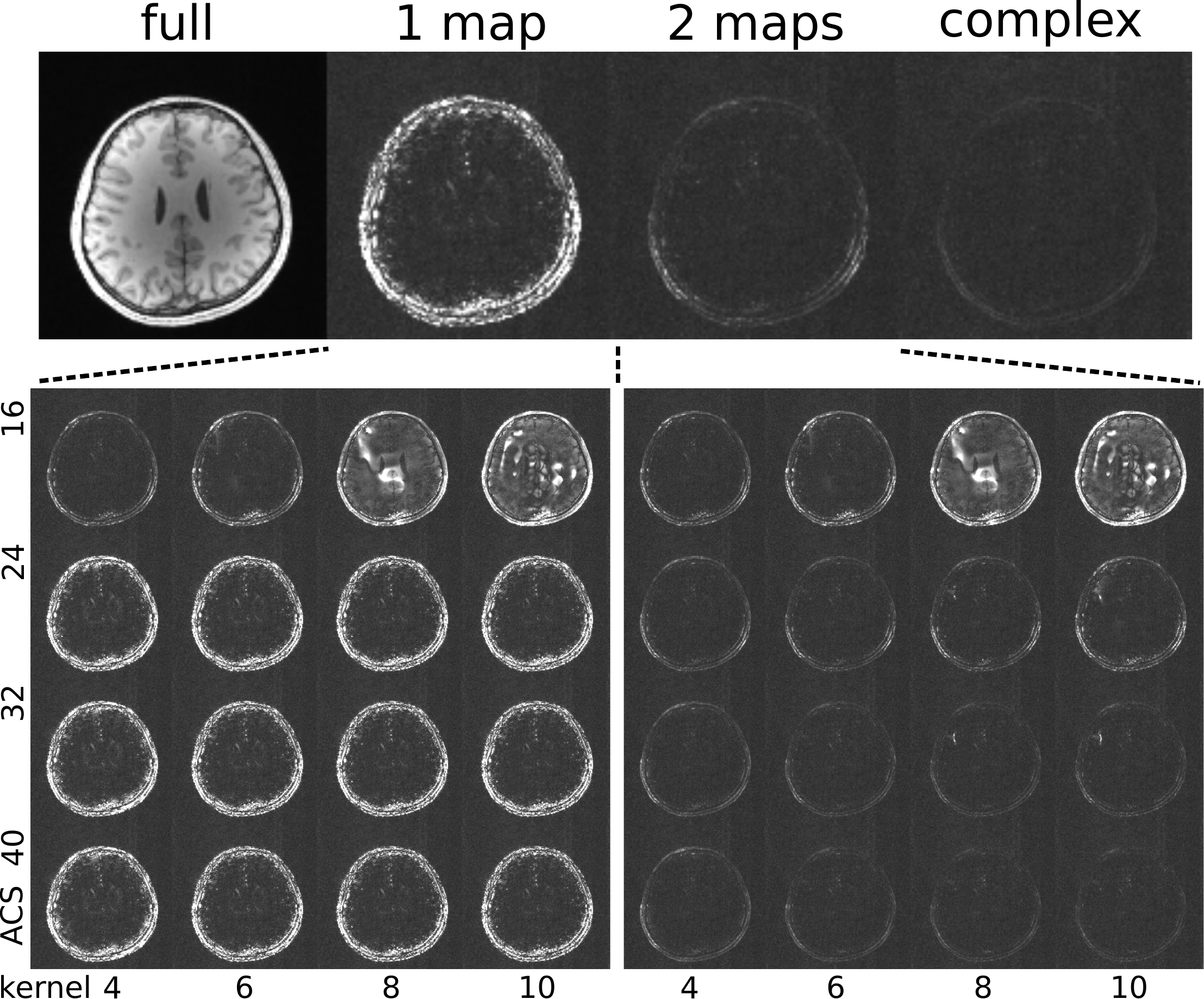}
\caption{\label{FIG3}
This figure shows the combined residual (unexplained) signal from
all channels after projection of the coil images onto different spaces.
{\bf Top:} 
full signal for comparison (full),
real-projection onto the first set of VCC-ESPIRiT maps (1 map), 
real-projection onto both sets of VCC-ESPIRiT maps (2 maps), 
complex-projection onto a single set of conventional ESPIRiT maps (complex).
{\bf Bottom:} The impact of the size of the calibration region and the size
of the kernel on the quality of the maps is evaluated using a real-projection onto 
one (left)
or both (right) sets using a calibration region (ACS) of size 16x16, 24x24, 32x32,
and 40x40 and kernel size of 4x4, 6x6, 8x8, and 10x10. 
For each case, the residual signals for all channels have been combined into 
a single image using the root-sum-of-squares method. Relative to the
full signal all other images have been scaled up by a factor of five
to aid visualization.
}
\end{figure}

Figure~\ref{FIG3} shows the results of the projection test
for the VCC-ESPIRiT maps.
The real projection $P_R$ of the coil images onto
the space spanned by the coil sensitivities shows
that the image and coil phase are accurately captured in most
parts of the image, but that residual signal occurs in areas with 
high-frequency phase from fat and blood vessels which cannot be
modelled with a single set of smooth maps even with a larger 
calibration region. With a second set of maps a
larger space is spanned and the residual
signal can be modelled. Although the quality of the maps
degrades if the calibration region is smaller than 24x24 and 
especially when a large kernel size is used together with a 
small calibration region, the quality of the single set of
VCC-ESPIRiT maps is  otherwise relatively robust to the choice of parameters.
When using just a single set of maps, increasing the calibration region and the kernel size does not improve the error much.
This is because high frequency phase can not be represented well without going to extreme sizes. In contrast, 
increasing the calibration region and kernel sizes does reduce the error when two sets of maps are used. In our example, for 40x40 calibration region using two sets of maps, while enforcing real-valued result, has comparable error to the case where phase is not constrained. This means that for this case,  VCC-ESPIRiT captures well the signal subspace with  all its phase variations.

\begin{figure}[h]
\includegraphics[width=\textwidth]{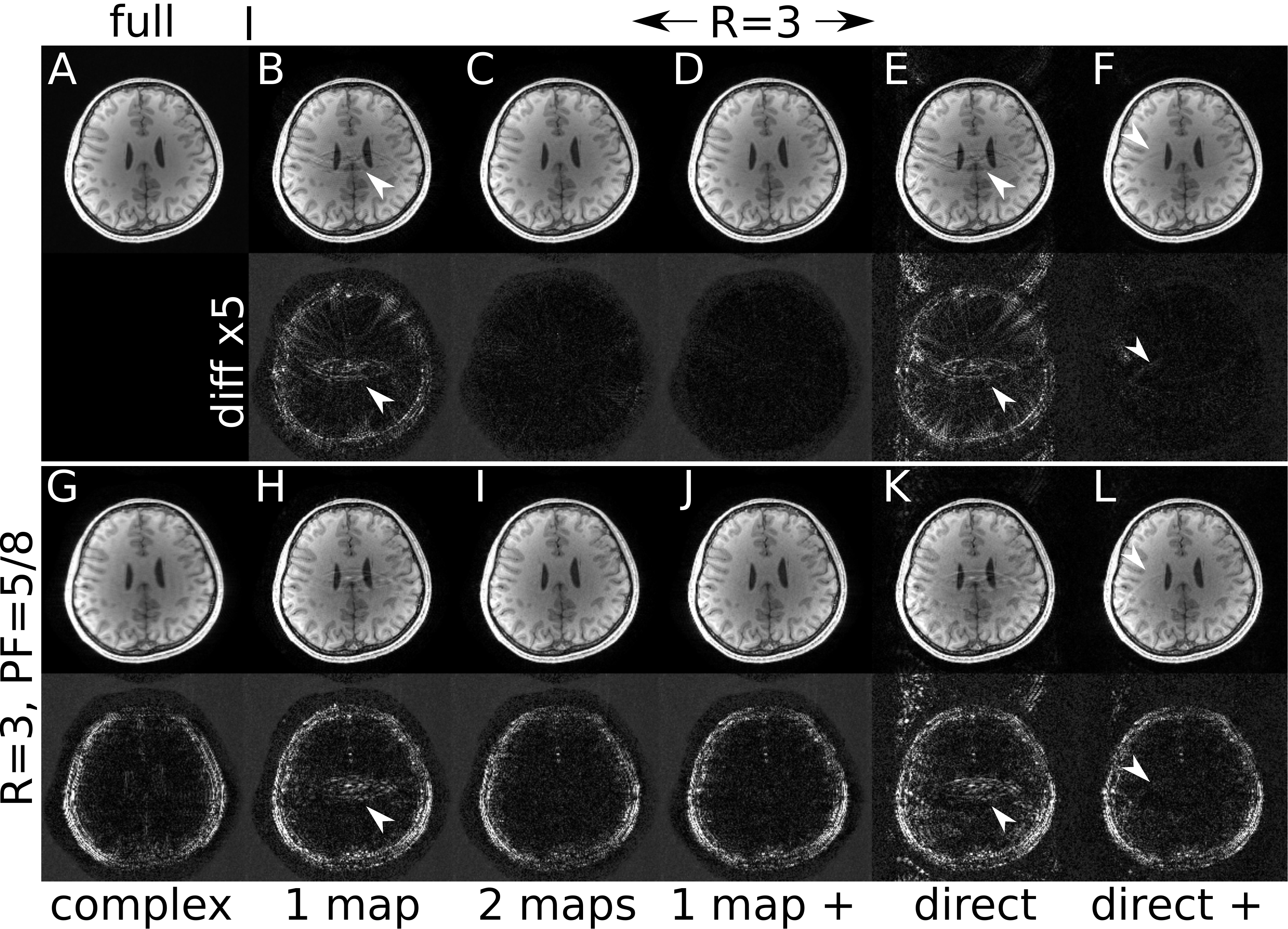}
\caption{\label{FIG4}
Reconstruction from fully-sampled data (A) and
iterative parallel imaging reconstruction (R = 3, 24x24 ACS lines)
with (top, B-F) and without (bottom, G-L) additional 
partial-Fourier sampling (PF=5/8).
Sensitivities were estimated with VCC-ESPIRiT using 
one set of maps (1 map, B,H,D,J) or two set of maps (2 maps, C,I) 
or directly from the k-space center (direct, E,F,K,L). Some
variants  use regularization of the imaginary 
component (indicated by $+$, D,J,F,L) or no phase constraint (complex, G).
If the high-frequency phase of the image is not correctly
modelled, a real-value constraint causes artifacts (arrows).
Difference images to the fully-sampled reference
are scaled up by a factor of five.}
\end{figure}

Figure~\ref{FIG4} demonstrates reconstruction of data from 
an accelerated parallel-imaging and a parallel-imaging
partial-Fourier acquisition using differently estimated maps
and various constraints. When using a single set of 
maps together with a real-value constraint, aliasing
artifacts are not completely resolved in the reconstruction
(B,E,F,H,K,L). By using a second set of VCC-ESPIRiT maps the real-value 
constraint is relaxed in image regions affected by high-frequency phase.
In this example, this yields reconstructions almost without visible
artifacts (C,I). Good reconstructions can also be obtained
when the real-value constraint is generally replaced by regularization 
of the imaginary component (D,J). Comparing the results using 
VCC-ESPIRiT maps with results using directly estimated maps (EF,KL)
confirms the better accuracy of ESPIRiT maps. Even
when using regularization of the imaginary part minor artifacts
remain in the case of directly estimated maps. For the partial
Fourier case, the iterative reconstruction without real-value 
constraint shows pronounced blurring due to the missing
information in one half of k-space (G). Using a real-value
constraint or regularization of the imaginary part
yields sharp images (H-L).

\begin{figure}[h]
\includegraphics[width=\textwidth]{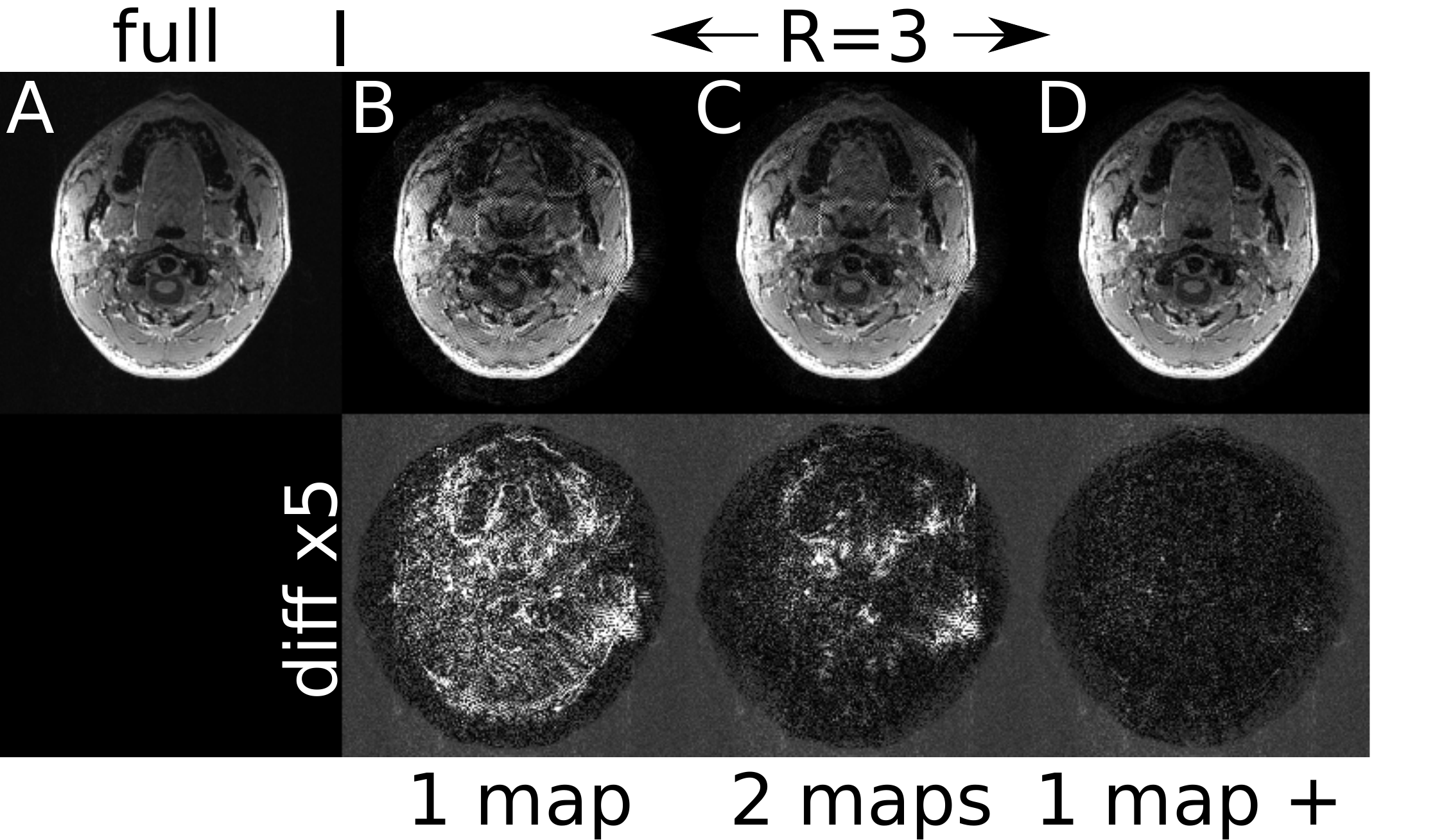}
\caption{\label{FIG5}
Reconstruction from fully-sampled data (full, A), VCC-ESPIRiT using one 
set of maps (1 map, B), VCC-ESPIRiT using two set of maps (2 maps, C), and
VCC-ESPIRiT using  one set of maps with regularization of the imaginary 
component (1 map +, D). 
Difference images to the fully-sampled reference
are scaled up by a factor of five.
}
\end{figure}

In Figure~\ref{FIG5} reconstructed images for a
slice with more phase variations are shown. Although 
the reconstruction using two sets of VCC-ESPIRiT maps
has significantly reduced errors compared to the 
reconstruction using only one set of maps, it still
fails to recover a good image due to unaccounted phase.
The combination of a single set of VCC-ESPIRiT maps
with regularization of the imaginary component yields
a good reconstruction.

\section{Discussion}

This work describes VCC-ESPIRiT, an extension to ESPIRiT for 
the computation of sensitivity maps which include image phase.
It combines ESPIRiT applied to virtual conjugate channels
with a post-processing step to determine the phase of the maps.
Having explicit sensitivities with image phase is
useful in a wide variety of image reconstruction tasks. In
phase-constrained reconstruction it can be used to exploit
conjugate symmetry to increase SNR or recover information
from partial-Fourier acquisitions. In model-based
image reconstruction the phase of the image must often be taken into 
account in the forward model.

Especially when only a limited amount of calibration data
is available, ESPIRiT maps are often more accurate than maps 
directly estimated from the k-space center~\cite{ESPIRiT}.
Still, even with ESPIRiT maps phase-constrained parallel imaging
with a single set of maps is not robust to errors from high-frequency 
phase variations which cannot be described by a smooth map. 
It has been shown previously that k-space methods based on 
virtual conjugate coils are more robust~\cite{Blaimer2015}.
Using the ESPIRiT formalism this observation can be explained:
In regions of the image with high-frequency phase variations
the reconstruction operator has a second eigenvalue close to
the value one. The corresponding eigenvector map corresponds to a second 
set of maps that is needed to span the data - 
indicating the location where the imaginary component of the signal 
cannot be neglected. As shown, using the second set of maps in a real-value 
constrained reconstruction is more robust, because it relaxes the
constraint at the corresponding locations to allow arbitrary complex 
values. Interestingly, this works because
the approximate location of small objects with high-frequency
phase such as blood vessels  can be recovered from a relatively small 
amount of data from the k-space  center. This property can be explained
by the relationship of ESPIRiT to classical subspace-based frequency-estimation 
methods~\cite{MUSIC,ESPRIT}.

As long as only high-frequency phase variations but no other
kinds of corruption occur, the essential information
are the locations where the real-value constraint has to
be relaxed. Although the real-value constraint could be modified
directly, using multiple sensitivity maps has the 
advantage of being robust also against other types of errors.
It should be noted that with other types of errors 
the multiple sets of maps might appear as an arbitrary linear 
mixture of eigenvectors to eigenvalue one~\cite{ESPIRiT,Bahri2014}.
If this is the case, the maps are usually not smooth due 
to mixing of different components. If smooth maps 
are desired, {\it e.g.} when sparsity constraints are applied
to the coil-combined image, additional alignment steps 
similar to the ones proposed for ESPIRiT-based coil 
compression are needed~\cite{Bahri2014}.

For images with a large amount of phase variations using
two ESPIRiT maps estimated from low-resolution calibration
data still does not sufficiently relax the constraints
for an artifact-free reconstruction.
In this case, it is necessary to replace the hard phase
constraint with an increased regularization of the
imaginary component, {\it i.e.} softening the constraint 
everywhere~\cite{Hoge07}. Although
the information from the second map is then not required,
this method still critically depends on 
accurate phase information and benefits from the
higher accuracy of the sensitivities estimated with ESPIRiT.
As another enhancement, a sparsity penalty applied directly
to the imaginary component may allow
further improvements~\cite{Li2015}.

If not enough calibration information is available, another 
reconstruction method can be used to first recover a larger 
calibration area from partially
undersampled k-space center, for example using iterative 
GRAPPA~\cite{Blaimer2015} or (robust) SAKE~\cite{Shin14,Zhu14}.
To exploit correlations between conjugate-symmetric parts of 
k-space at this stage, this could be applied to
a k-space already extended with virtual conjugate coils,
{\it i.e.} using VCC-SAKE. In fact, a very similar method
called LORAKS has recently been proposed~\cite{LORAKS}.

Finally, if the phase is of interest in itself, {\it i.e.} in 
phase-contrast imaging, a possible approach is to synthesize 
regular coil images from the reconstructed image  by multiplying 
with the sensitivities. The synthetic coil images then 
include all phase information and can be further processed
as with other methods. Ofcourse, the use of any type of
phase contraint must be  considered very carefully in these
applications.

\section{Conclusion}

An extension to ESPIRiT has been presented, which can be used to estimate
coil sensitivities which include slowly-varying image phase. 
The sensitivities with phase can be directly used in a reconstruction with 
real-value constraint to improve SNR or to exploit conjugate-symmetry 
in k-space for partial Fourier acquisitions. In case of high-frequency
phase variations, 
VCC-ESPIRiT yields  a second set of maps in affected regions, 
which can then be taken into account to make phase-constrained reconstructions 
more robust. If the image has a large amount of high-frequency phase,
robust reconstruction requires the phase constraint to be relaxed.

\section*{Acknowledgements}

The authors thank Mariya Doneva for valuable comments.

\bibliography{phase}

\begin{thebibliography}{10}

\bibitem{Bahri2014}
D.~Bahri, M.~Uecker, and M.~Lustig.
\newblock Three-nearest-neighbor alignment for smooth {ESPIRiT} maps.
\newblock In {\em Proceedings of the 22nd Annual Meeting ISMRM}, page 4394,
  Milano, 2014.

\bibitem{VCGRAPPA}
M.~Blaimer, M.~Gutberlet, P.~Kellman, F.~A. Breuer, H.~K\"ostler, and M.~A.
  Griswold.
\newblock Virtual coil concept for improved parallel {MRI} employing conjugate
  symmetric signals.
\newblock {\em Magn\ Reson\ Med}, 61:93--102, 2009.

\bibitem{Blaimer2015}
M.~Blaimer, M.~Heim, D.~Neumann, P.~M. Jakob, S.~Kannengiesser, and F.~A.
  Breuer.
\newblock Comparison of phase-constrained parallel {MRI} approaches: Analogies
  and differences.
\newblock {\em Magn\ Reson\ Med}, 2015.
\newblock Epub, doi: 10.1002/mrm.25685.

\bibitem{LORAKS}
J.~P. Haldar.
\newblock Low-rank modeling of local $k$-space neighborhoods {(LORAKS)} for
  constrained {MRI}.
\newblock {\em IEEE Trans\ Med\ Imaging}, 33:668--681, 2014.

\bibitem{Hoge07}
W.~S. Hoge, M.~E. Kilmer, C.~Zacarias-Almarcha, and D.~H. Brooks.
\newblock Fast regularized reconstruction of non-uniformly subsampled
  partial-fourier parallel mri data.
\newblock In {\em Biomedical Imaging: From Nano to Macro (ISBI). 4th IEEE
  International Symposium on}, pages 1012--1015, Washington, 2007.

\bibitem{Inati2013}
S.~J. Inati, M.~S. Hansen, and P.~Kellman.
\newblock A solution to the phase problem in adaptive coil combination.
\newblock In {\em Proceedings of the ISMRM 21th Annual Meeting}, page 2672,
  Salt Lake City, 2013.

\bibitem{Li2015}
G.~Li, J.~Hennig, E.~Raithel, M.~B\"uchert, D.~Paul, J.~G. Korvink, and
  M.~Zaitsev.
\newblock An {L1}-norm phase constraint for half-{Fourier} compressed sensing
  in {3D} {MR} imaging.
\newblock {\em Magnetic Resonance Materials in Physics, Biology and Medicine},
  28:459--472, 2015.

\bibitem{Margosian1986}
P.~Margosian, F.~Schmitt, and D.~Purdy.
\newblock Faster {MR} imaging: imaging with half the data.
\newblock {\em Health Care Instrum}, 1:195, 1986.

\bibitem{McKenzie2002}
C.A. McKenzie, E.N. Yeh, M.A. Ohliger, M.D. Price, and D.K. Sodickson.
\newblock Self-calibrating parallel imaging with automatic coil sensitivity
  extraction.
\newblock {\em Magn\ Reson\ Med}, 47:529--538, 2002.

\bibitem{HOMODYNE}
D.~C. Noll, D.~G. Nishimura, and A.~Macovski.
\newblock Homodyne detection in magnetic resonance imaging.
\newblock {\em IEEE Trans\ Med\ Imaging}, 10:154--163, 1991.

\bibitem{PruessmannSENSE}
K.~P. Pruessmann, M.~Weiger, M.~B. Scheidegger, and P.~Boesiger.
\newblock {SENSE}: Sensitivity encoding for fast {MRI}.
\newblock {\em Magn\ Reson\ Med}, 42:952--962, 1999.

\bibitem{ESPRIT}
R.~Roy and T.~Kailath.
\newblock {ESPRIT} - {E}stimation of signal parameters via rotational
  invariance techniques.
\newblock {\em IEEE Trans Acoust Speech Signal Process}, 37:984--995, 1989.

\bibitem{POCSENSE}
A.~A. Samsonov, E.~G. Kholmovski, D.~L. Parker, and C.~R. Johnson.
\newblock {POCSENSE}: {POCS}-based reconstruction for sensitivity encoded
  magnetic resonance imaging.
\newblock {\em Magn\ Reson\ Med}, 52:1397--1406, 2004.

\bibitem{MUSIC}
R.~O. Schmidt.
\newblock Multiple emitter location and signal parameter estimation.
\newblock {\em IEEE Trans Antennas Propag}, 34:276--280, 1986.

\bibitem{Shin14}
P.~J. Shin, P.~E.~Z. Larson, M.~A. Ohliger, M.~Elad, J.~M. Pauly, D.~B.
  Vigneron, and M.~Lustig.
\newblock Calibrationless parallel imaging reconstruction based on structured
  low-rank matrix completion.
\newblock {\em Magn\ Reson\ Med}, 72:959--970, 2014.

\bibitem{UeckerSeg}
M.~Uecker, A.~Karaus, and J.~Frahm.
\newblock Inverse reconstruction method for segmented multishot
  diffusion-weighted {MRI} with multiple coils.
\newblock {\em Magn\ Reson\ Med}, 62:1342--1348, 2009.

\bibitem{ESPIRiT}
M.~Uecker, P.~Lai, M.~J. Murphy, P.~Virtue, M.~Elad, J.~M. Pauly, S.~S.
  Vasanawala, and M.~Lustig.
\newblock {ESPIRiT} -- an eigenvalue approach to auto-calibrating parallel
  {MRI}: Where {SENSE} meets {GRAPPA}.
\newblock {\em Magn\ Reson\ Med}, 71:990--1001, 2014.

\bibitem{UeckerMilano14}
M.~Uecker and M.~Lustig.
\newblock Robust partial {Fourier} parallel imaging using {ESPIRiT} and virtual
  conjugate coils.
\newblock In {\em Proceecings of the 22nd Annual Meeting ISMRM}, volume~22,
  page 1629, Milano, 2014.

\bibitem{UeckerToronto15}
M.~Uecker, F.~Ong, J.~I. Tamir, D.~Bahri, P.~Virtue, J.~Y. Cheng, T.~Zhang, and
  M.~Lustig.
\newblock Berkeley advanced reconstruction toolbox.
\newblock In {\em Proceedings of the 23rd Annual Meeting ISMRM}, volume~23,
  page 2486, Toronto, 2015.

\bibitem{UeckerSaltLake13}
M.~Uecker, P.~Virtue, S.~S. Vasanawala, and M.~Lustig.
\newblock {ESPIRiT} reconstruction using soft {SENSE}.
\newblock In {\em Proceedings of the 21st Annual Meeting ISMRM}, volume~21,
  page 127, Salt Lake City, 2013.

\bibitem{Walsh2000}
D.~O. Walsh, A.~F. Gmitro, and M.~W. Marcellin.
\newblock Adaptive reconstruction of phased array {MR} imagery.
\newblock {\em Magn\ Reson\ Med}, 43:682--690, 2000.

\bibitem{Willig2005}
J.~D. {{Willig-Onwuachi}}, E.~N. Yeh, A.~K. Grant, M.~A. Ohliger, C.~A.
  McKenzie, and D.~K. Sodickson.
\newblock Phase-constrained parallel {MR} image reconstruction.
\newblock {\em J Magn Reson}, 176:187--198, 2005.

\bibitem{Zhu14}
D.~Zhu, M.~Uecker, J.~Y. Cheng, Z.~Bi, K.~Ying, and M.~Lustig.
\newblock Calibration for parallel {MRI} using robust low-rank matrix
  completion.
\newblock In {\em Proceedings of the 22nd Annual Meeting ISMRM}, volume~22,
  page 741, Milano, 2014.

\end{thebibliography}

\end{document}